\documentclass[10pt,twocolumn,letterpaper]{article}

\usepackage[]{wacv} % To force page numbers, e.g. for an arXiv version

\usepackage{graphicx}
\usepackage{amsmath}
\usepackage{amssymb}
\usepackage{booktabs}

\usepackage[pagebackref,breaklinks,colorlinks]{hyperref}

\usepackage[capitalize]{cleveref}
\crefname{section}{Sec.}{Secs.}
\Crefname{section}{Section}{Sections}
\Crefname{table}{Table}{Tables}
\crefname{table}{Tab.}{Tabs.}

\def\wacvPaperID{1425} % *** Enter the WACV Paper ID here
\def\confName{WACV}
\def\confYear{2025}

\usepackage{tikz}
 
\usepackage{amssymb,pifont,xcolor}

\begin{document}

%%%%%%%%% TITLE - PLEASE UPDATE
\title{CM3T: Framework for Efficient Multimodal Learning for Inhomogeneous Interaction Datasets}

\author{Tanay Agrawal\\
\normalsize INRIA\\
\normalsize Sophia Antipolis, France\\
\footnotesize\tt tanay.agrawal@inria.fr
% For a paper whose authors are all at the same institution,
% omit the following lines up until the closing ``}''.
% Additional authors and addresses can be added with ``\and'',
% just like the second author.
% To save space, use either the email address or home page, not both
\and
Mohammed Guermal\\
\normalsize INRIA\\
\normalsize Sophia Antipolis, France\\
%\footnotesize\tt mohammed.guermal@inria.fr
\and
Michal Balazia\\
\normalsize INRIA\\
\normalsize Sophia Antipolis, France\\
%\footnotesize\tt michal.balazia@inria.fr
\and
Francois Bremond\\
\normalsize INRIA\\
\normalsize Sophia Antipolis, France\\
%\footnotesize\tt francois.bremond@inria.fr
}

\maketitle

%%%%%%%%% ABSTRACT
\begin{abstract}
Challenges in cross-learning involve inhomogeneous or even inadequate amount of training data and lack of resources for retraining large pretrained models. Inspired by transfer learning techniques in NLP, adapters and prefix tuning, this paper presents a new model-agnostic plugin architecture for cross-learning, called CM3T, that adapts transformer-based models to new or missing information. We introduce two adapter blocks: multi-head vision adapters for transfer learning and cross-attention adapters for multimodal learning. Training becomes substantially efficient as the backbone and other plugins do not need to be finetuned along with these additions. Comparative and ablation studies on three datasets Epic-Kitchens-100, MPIIGroupInteraction and UDIVA v0.5 show efficacy of this framework on different recording settings and tasks. With only 12.8\% trainable parameters compared to the backbone to process video input and only 22.3\% trainable parameters for two additional modalities, we achieve comparable and even better results than the state-of-the-art. CM3T has no specific requirements for training or pretraining and is a step towards bridging the gap between a general model and specific practical applications of video classification.
\end{abstract}

%%%%%%%%% BODY TEXT
\section{Introduction}
\begin{figure}[ht]
\begin{center}
\includegraphics[width=0.4\textwidth]{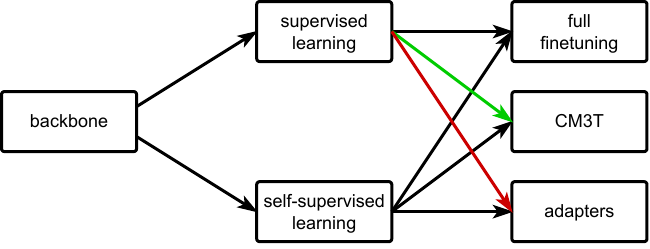}
\end{center}
\caption{This is a representation of the main problem CM3T aims to solve. Backbones pretrained using self-supervised learning provide good general features, thus all methods of finetuning work well. In the case of supervised pretraining, adapters fail to perform well (in red) and CM3T is introduced to solve this (in green).}
\label{figure1}
\end{figure}

% Paragraph establishes the background of the work - inhomogenous data, parameter efficient transfer learning.

Video classification is a big field in computer vision with various sub-tasks and datasets for each of these tasks. Recently, there has been an increase in tasks, datasets, and recorded modalities. Most work is specific to a task with corresponding datasets or a subset of these modalities, and their modification for a new input protocol is tedious.
% There is also a lot of existing overlapping work that could be combined to achieve good results.
Methods including late and early fusion and cross-attention are generally used for combining them, but they are not the most efficient way to treat this wide variety of data.
Thus, there is a need for a method that can handle this increase in data having high variability in structure and which learns robust relations that are shareable among tasks and datasets. The field of parameter efficient transfer learning~(PETL) is increasing in popularity to answer this problem. The basic idea consists in adding adapters\footnote{In this paper, we refer to \textit{adapters} including a mix of multiple techniques as in the M\&M adapters.} (i.e., plugin architectures of very few trainable parameters) to a backbone and only train these while keeping the backbone frozen. With increasing model and dataset sizes, PETL techniques facilitate finetuning only adapters with less resources and time compared to full-finetuning (i.e., backbones + adapters).

The video backbones used as a starting point for PETL can be pretrained using either i) the traditional supervised method on big datasets or using ii) more sophisticated self-supervised methods which result in better general features, such as VideoMAE~\cite{videomae} or contrastive learning such as CLIP~\cite{clip}. Existing PETL techniques only work well after using the latter (i.e., self-supervised pretrained backbones). 
But, self-supervised pretrained backbones
% \footnote{For ease of understanding, we refer to specific pretraining requirements as self-supervised in this paper. For example, this also includes CLIP which is based on contrastive learning.} 
are not widely available for use off the shelf and their training is resource intensive. For example, dual-path adapters~\cite{dual_path} and ST-adapters~\cite{st_adapter} require a backbone pretrained with CLIP. However, most works on self-supervised pretraining methods only use Vit/ViViT. Swin/Video-Swin transformers have not been pretrained using these self-supervised methods despite their superior performance. The main motivation behind this work is to propose new adapters to work well with traditional supervised pretrained backbones. Figure~\ref{figure1} summarizes this, the red arrow signifies the problem we are trying to solve and the solution is in green.

%But, if a specific pretraining step is necessary, for example Dual path adapters~\cite{dual_path} and ST Adapters~\cite{st_adapter} require a backbone pretrained using CLIP, it defeats this purpose.
% Available foundation models which are pretrained using self-supervised methods are not available for many datasets and tasks, and need to be either finetuned to achieve the best results or an additional step for meeting the pretraining requirements of exisiting methods. This requires a lot of time and resources and CM3T addresses this problem.

% Paragraph discusses basic technical contributions

We introduce CM3T (Cross Multimodal Multi-dataset Multitask Transformer), a novel PETL technique, which can leverage these new adapters.
%is able to perform well with plain supervised pretrained backbones. 
CM3T takes a frozen backbone, for example, the Video Swin Transformer~\cite{swin} pretrained (i.e., fully fine-tuned) on Kinetics-400 or Something-Something v2, and adds plugins (i.e., adapters) in parallel without changing the backbone architecture. Only these plugins need to be trained for downstream tasks and different datasets. Inspired by the Mix-and-Match~(M\&M) adapters~\cite{1313}, we combine prefix tuning with a newly introduced plugin, multi-head vision adapters. These adapters (shown in blue in Figure~\ref{figure2}.) improve upon existing scaled parallel adapters
% which add a set of bottleneck linear layers in parallel to modules in transformers. Simplified, they are 
by separating the processing for different spatial chunks into different heads of the input. This greatly increases performance as interaction datasets generally have almost fixed cameras and various objects and parts of the body always occur in particular spatial locations which generally remain the same. 
In addition, an approximation for prefix tuning, which has been proven to work well, is used as done by~\cite{13he2021towards}, but with some modifications. This is shown in red in Figure~\ref{figure2}. The details are discussed in Section~\ref{section3}.

% Paragraph discusses the next contribution - cross-attention adapters.

% Recently, most datasets have multimodal information and this work leverages the above idea to utilize this additional information.
Furthermore, the above idea can be further extended to cross-modal learning where the weights of the pretrained model do not have to be changed to incorporate new modalities, just as the backbone doesn't have to be changed to adapt to new datasets. This facilitates the use of existing work to build more complex systems. For this, we introduce the third and final module in Figure~\ref{figure2}, called cross-attention adapters (in green) for multimodal learning. 
Since cross-attention has been established as an effective manner for multimodal learning, we show how to incorporate it in place of linear layers in adapters, allowing their use for multimodal learning as well. It allows CM3T to learn the relationships between vision and other modalities while retaining its other advantages. This is a challenging task to execute in a resource efficient manner as increasing the number of input modalities generally increases the number of branches, hence the resources used. But, the theory of adapters allow us to overcome this. Thus, this contribution is significant as shown by the results in Section~\ref{section4}.
Challenges in processing multimodal data include heterogeneity of the present modalities, lack of correlation between modalities (for example different pitches in the audio could correspond to the same text), and the need for many training samples for convergence. Building upon each challenge above in order, CM3T addresses these challenges with the following additions. 
% making the architecture modular, and sharing weights across tasks and datasets.
Adding a new modality is cumbersome as it requires retraining parts of the backbone along with the new branches for the modality itself, but with this framework, it would just be a new plugin which is trainable by itself.
To capture the relationship between different modalities, we add an additional module to capture the relationships between all modalities other than vision (the backbone), when available. 
To make training faster and convergence easier as compared to using the generic embedding from large transformer models, the downsampling layer in adapters provides a good embedding to use for cross-attention.
Additionally, training cross-modal adapters across datasets improves performance and provides a good pretrained feature extractor for small datasets.

% Breif description of experiments and results.

To show that CM3T is suitable for multimodal, multi-dataset and multitask learning, we experiment on three different datasets with different recording scenarios and tasks: Epic-Kitchens-100~(EK-100) with first-person human-object interaction videos, MPIIGroupInteraction~(MPIIGI) and UDIVA~v0.5~(UDIVA) with human-human interactions in group settings while talking or doing different tasks respectively. We choose a mix of small and large multimodal interaction datasets to show the efficacy of our work in different settings. We show that we achieve comparable accuracy to state-of-the-art for all the datasets using only 12.8\% trainable parameters as compared to the backbone to process video input and only 22.3\% trainable parameters to process two additional modalities. We perform additional experiments to study how CM3T works in different scenarios and explore the reasons for the results obtained.

% (*) Contributions:
In summary, 
%the summary of 
our contributions are: 
\begin{itemize}
\item We introduce multi-head vision adapters which perform well with traditional supervised pretraining, in contrast to existing PETL techniques.
\item We introduce cross-attention adapters which are easier to modify than traditional multimodal methods. They also benefit from weight sharing, similarly to traditional adapters, by storing the relations between vision and other modalities and reusing them later.
\item We provide a framework, CM3T, for combining these techniques along with an approximation of prefix tuning to achieve state-of-the-art performance.
\end{itemize}

% \begin{quote}\begin{scriptsize}\begin{verbatim}
% \documentclass[letterpaper]{article}
% \usepackage[submission]{aaai24}
% \end{verbatim}\end{scriptsize}\end{quote}

\section{Related Work}

\subsection{Parameter Efficient Task Adaptation}
% PETL is established on one basic principle which is described below. 
Transformer-based backbones, such as Video Swin Transformer~\cite{swin} or ViVit~\cite{vivit}, are state-of-the-art feature extractors which are carefully trained on big datasets using either supervised or the better performing self-supervised methods. But finetuning these models is resource-intensive and does not converge for small datasets. The  main theory behind all PETL work for computer vision is that finetuning any general feature extractor involves learning the environment in which the new data is recorded and the intricacies of the new task. The basic spatial understanding of the video remains the same. Thus, we can use this basic understanding by these pretrained models and employ only a few additional parameters to learn the new information.
% We take a frozen pretrained backbone (Video Swin Transformer~\cite{swin}) and add existing and newly introduced plugins in parallel without changing the backbone and train only these plugins for downstream tasks and different datasets.

The field of NLP has seen a lot of work following the above idea, such as adapters~\cite{adapters}, LoRA~\cite{lora}, and prefix tuning~\cite{28li2021prefix}. These methods get similar results while adding less than 10\% parameters to existing models which are trained to learn the new task while the pretrained weights are frozen. These have also been extended to computer vision~\cite{vl_adapters,AF,VPT,pruning_adapters}.

There are three recent PETL methods which show good results: (1)~only updating new parameters added to the model or the input~\cite{adapters,hyperformer,p24,28li2021prefix}; (2)~updating some of the parameters of the model in a sparse manner~\cite{12zong2021proceedings,47sung2021training,52zaken2021bitfit}; and (3)~low-rank factorization of weight matrices to reduce the number of parameters to be updated while keeping the weight matrix approximately the same~\cite{17hu2021lora}. Combining these approaches, \cite{13he2021towards,34mao2021unipelt} propose a unified parameter efficient training framework. Among these approaches, adapters, which belong to the first category, have been used in computer vision~\cite{41rebuffi2017learning,42Rebuffi_2018_CVPR} and natural language processing~\cite{adapters,20mahabadi2021parameter,33karimi2021compacter}. While adapters add more parameters into models, prompt-based approaches instead add trainable parameters to inputs~\cite{11gu2021ppt,p24,28li2021prefix}, and experiments have shown their value in language and vision tasks. We use both techniques in~\cite{13he2021towards} as an inspiration for CM3T. VL Adapters~\cite{vl_adapters} compare various adapter techniques~\cite{adapters,hyperformer,33karimi2021compacter} applied to question answering tasks, but not to pure vision tasks. Their work aims to use adapters to project vision and language pretrained model embeddings into the language model's space whereas we show that it is possible to do it across vision datasets and also be used to add new modalities.

AdaptFormer~\cite{AF} uses adapters with only the linear layers of a transformer and achieves better results than full finetuning. But it uses VideoMAE~\cite{VM} for pretraining ViT~\cite{ViT} which is not feasible if resources are limited and cannot be used to make a generalized framework. Their method fails with models not carefully pretrained using self-supervised methods. Similarly, ST-adapters~\cite{st_adapter} use ViT pretrained using CLIP. They convert image models to video models using convolutions for time aggregation in addition to the upsampling and downsampling linear layers in an adapter and it works well, except for the case when traditional supervised pretraining is employed. Visual prompt tuning~(VPT)~\cite{VPT} uses prompt tuning for images, but prompts alone do not work well for videos which is also mentioned by~\cite{AF}.

The paper~\cite{pruning_adapters} shows that adapters only work for vision if the bottleneck dimension is large. They introduce a pruning technique to reduce the size of these adapters. We introduce multi-head vision adapters as an alternative that works well even with a small bottleneck dimension and without any specific pretraining method. Dual-path adaptation from image to video transformers~\cite{dual_path} show better results compared to others using supervised training methods, but it is still not comparable to full finetuning. They also have a specific input method that limits the maximum temporal size of input that can be provide which makes their model less scalable and not suitable for all datasets and downstream tasks.

\subsection{Multimodal Learning}
There is an inherent difference between videos and other modalities, such as audio or text, and thus it is challenging to combine them into one model. VATT~\cite{vatt} uses early fusion, where they concatenate all input modalities. Although the earlier the fusion, the better the results, there is a trade-off with the amount of data required for training as it is harder for models with early stage fusion to converge which leads to tedious self-supervised learning.

Some works design a specialized architecture for fusion at feature level~\cite{Visapp_paper,udiva}. These work better but there are limitations as the fusion is done after downsampling the input features which leads to loss of information and poor cross-modality relations. \cite{delbrouck-etal-2020-transformer,DBLP:journals/corr/abs-2003-01043,multilogue} have feature level fusion with minimal downsampling, but lack in handling specific modalities differently. So, there is a need for a model which can benefit from cross-modality learning at different levels. To answer this and so make the model flexible, we propose using cross-attention added to each block of a transformer architecture. State-of-the-art methods M\&M Mix~\cite{1313} and MuMu~\cite{MuMu} are either modality specific or have a rigid architecture making it hard to add and remove modalities. This work addresses these drawbacks by having a flexible architecture that can accommodate any type of input.

\section{CM3T Framework}
\label{section3}

\begin{figure*}[ht]
\begin{center}
\includegraphics[width=0.9\textwidth]{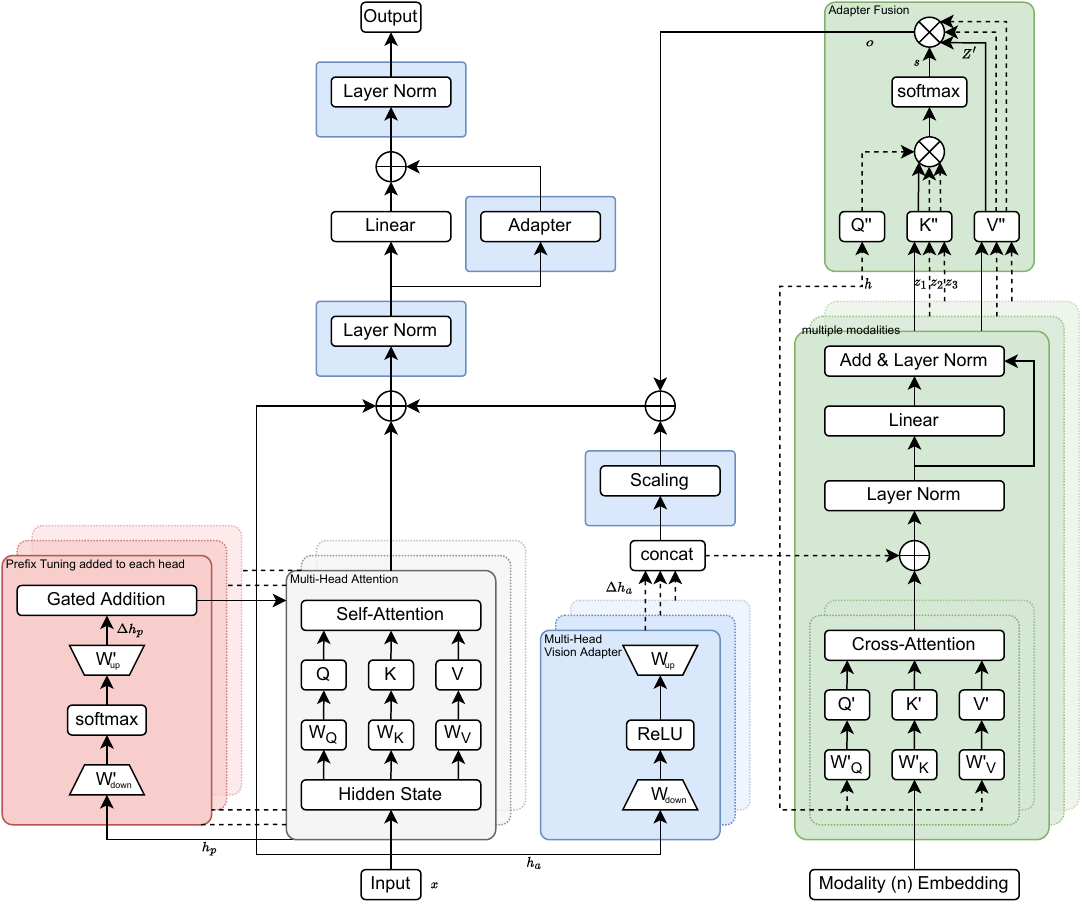}
\end{center}
\caption{Detailed architecture of CM3T. Colored parts are the ones that are finetuned and the rest are frozen. It has three separate blocks added to it which are shown in three different colors. Prefix tuning is complicated to show in detail, so only a schematic is shown. 
% Equations~\ref{eq_2} and~\ref{eq_4} show how the upscaling and downscaling weights are computed.
% Equations~\ref{eq_2} and~\ref{eq_4} being similar shows how prefix tuning is approximated as adapters.
The rest of the details are described in Section~\ref{section3}.}
\label{figure2}
\end{figure*}

We define an easy way to use existing multimodal data and pretrained models when approaching a video classification or video understanding task. This will assist in bridging the gap between research and practical applications. This section discusses some of the technical details of the background and then the methodology of our work.

\subsection{Choosing a Pretrained Model}
Our method is focused on transformer-based backbones which have produced state-of-the-art results for various vision tasks. We use the Video Swin Transformer~(Video Swin-B)~\cite{swin}, but the following steps of the framework are model invariant and the backbone can be chosen according to the need. The reason for choosing Video Swin-B is that different blocks process the input at different spatial resolutions. Depending on the side input (other modalities), cross-attention performs well with different blocks, that is, different spatial resolutions.

\subsection{Finetuning or Using Adapters}
Once we have a pretrained vision model, the next step is to finetune and adapt it to the target dataset. If computational resources or time are a constraint, adding adapters and prefix tuning and training them in place of full finetuning produces comparable results with significantly fewer parameters to train. There is also the possibility of combining this step with the following steps (in this section and the next one) for end-to-end learning, but we perform each step separately to compare their performance with corresponding state-of-the-art. The results for end-to-end training are also shown in the next section.

\subsubsection{Background}
We take inspiration from scaled parallel adapters and prefix tuning (PT) as used by~\cite{13he2021towards}. Figure~\ref{figure2} shows all the additions to the pretrained model along with our modifications. Multi-head vision adapters (MHVA) (in blue) and prefix tuning (in red) are discussed in this subsection and cross-attention adapters (CAA) (in green) are discussed in Section~\ref{multimodality}.

Mathematically, adapters from~\cite{adapters} are defined as
\begin{equation} \label{eq_adapter_basic}
y = s\cdot \Delta h_a
\end{equation}
\begin{equation} \label{eq_2}
\Delta h_a = \mathit{ReLU}\!(h_aW_\mathit{down}) \cdot W_\mathit{up}
\end{equation}
where $h_a = x$ is the input of size $d$, $W_\mathit{down}\in\mathbb{R}^{d \times r}$ is the weight matrix for the down-projection layer with bottleneck dimension $r$, $W_\mathit{up}\in\mathbb{R}^{r \times d}$ is the up-projection layer, and $s$ is the scaling factor. We use this in parallel instead of sequential, similar to~\cite{13he2021towards}. We also use their definition for prefix tuning (for simplification, Figure~\ref{figure2} does not show recurrent connection for prefix tuning),
\begin{equation} \label{eq_3}
    h_p \leftarrow (1-\lambda)\cdot h_p + \lambda\cdot \Delta h_p
\end{equation}
\begin{equation} \label{eq_4}
    \Delta h_p = \mathit{softmax}\!\left(h_pW_{down}^{'}\right) \cdot W_{up}^{'}
\end{equation}
\begin{equation}
    W_{down}^{'} = W_{q}P_{k}^{T} \qquad W_{up}^{'} = P_{v}
\end{equation}
where $W_{q}$ is the weight matrix for getting query vector from the input $h_{p}$, $P_k = C \cdot W_k$ and $P_v = C \cdot W_v$ are prefix tuning vectors which are learned using $W_k$ and $W_v$ (key and query weight matrices of the transformer backbone). Here, $C$ is a learned embedding which is randomly initialized and $\lambda$ is the factor used for gated addition. The red part of Figure~\ref{figure2} shows prefix tuning added to transformers, it is added in parallel to each head of multi-head attention.

\subsubsection{Incorporating Multi-Head Vision Adapter and Prefix Tuning into CM3T}
Using adapters for vision tasks is more challenging than NLP as language understanding does not change with the task or dataset, but video datasets have a wide variety of settings, such as indoor or outdoor recording scene, different views and camera angles, lighting changes, and more. Finetuning allows the networks to overcome these changes, but it is hard for adapters owing to less capability to change the original model's activations. But with a few changes, adapters can show performance comparable to fully finetuned models. Blue parts of Figure~\ref{figure2} mark the adapters.

AdaptFormer~\cite{AF} adds scaled parallel adapters to linear layers only and achieves better results than finetuning owing to a sophisticated pretrained ViT model using VideoMAE~\cite{VM}. We achieve very poor performance with the same method without this specific pretraining, even when coupled with prefix tuning. So, this leads to our first change, inspired by multi-head attention, we introduce \textbf{Multi-Head Vision Adapter~(MHVA)}. This is different from multi-head attention as the input is divided along the window dimension of Video Swin transformers (or spatial patch dimension in ViViT) and not the channel dimension. Essentially, there are different linear layers for different sets of windows/patches. We saw that increasing the bottleneck dimension in adapters only increased the performance slightly (as shown by~\cite{pruning_adapters}), but adding the above change allowed the network to learn better even with a smaller bottleneck dimension. Overall, the parameters do not increase by a big margin as compared to traditional adapters as we use a smaller bottleneck dimension. To define the change mathematically, the input $h$ is divided along the window dimension to get $\{h_1, h_2, h_3, \ldots\}$. Each has its own parallel adapter and the output is concatenated along the same dimension before scaling and addition. Extending Equation~\ref{eq_3},
\begin{equation}
\{h_{a1}, h_{a2}, \ldots\} \leftarrow \{h_{a1}, h_{a2}, \ldots\} + s\cdot \Delta \{h_{a1}, h_{a2}, \ldots\}
\end{equation}
where each operation is performed element-wise.

Our second change is that we make the scaling factor for adapters ($s$ in Equation~\ref{eq_adapter_basic}) added to linear layers learnable, allowing greater change to activations. Attention in pretrained models might focus on features that are not relevant to the new downstream task or dataset, but this change allows adapters to overcome this. For EK-100, when the value is fixed at $4.0$, we achieve 1.1\% lower performance.

Without the two changes mentioned above, adapters have very poor performance for the domain of computer vision with traditionally available pretrained models. These adapters are named multi-head vision adapters. These are specific to Video Swin transformers, but the same concept can be applied to modify adapters for any model using different linear layers in adapters for different sets of windows to which attention is applied. Section~\ref{section4} shows results for ViViT-B as a backbone model too. The reason for good performance with this addition is that it gives the adapters the ability to learn different representations for different chunks of the input.

The third change is more specific as compared to the first two. We use ReLU activation in place of tanH with a lower dropout for \textbf{Prefix Tuning~(PT)} and that provides a smoother training curve and easier convergence. It also allows for 0.7\% gain in accuracy on EK-100 dataset.

\subsection{Adding Other Modalities (CAA)}
\label{multimodality}
\textbf{Cross-attention adapters} are used for adding modalities to the model received from the previous step. Cross-attention adapters are simply obtained by replacing the two linear layers in the adapters with a cross-attention module. Each added modality has its own adapter. The query and value inputs to this adapter are taken from the concatenation of hidden states from the bottleneck hidden state in the multi-head vision adapter $Q = V = \mathit{ReLU}(xW_\mathit{down})$, where $h$ is the input to the Video Swin-B block and $W_\mathit{down}$ is the same as that in Equation~\ref{eq_2}. The key is taken as the feature embedding from the new modality.

To merge all the adapters trained for different modalities, in place of simple addition, \textbf{AdapterFusion}~\cite{AFu} is used which captures the interaction between different side inputs i.e, modalities other than vision. It is an attention block where each head has the same query as that of attention in cross-attention adapter for each modality, described above, let's say $h$. The key and value for each head are taken from the output $z_n$ of each cross-attention adapter with $n$ signifying the $n$-th modality. The module is expressed as
\begin{equation}
    s' = \mathit{softmax}\!\left(h^TW_Q \bigotimes z_n^TW_K\right)\mathrm{,} \, n\in\{1, \ldots, N\}
\end{equation}
\begin{equation}
    z'_n = z_nW_V \quad \mathrm{,} \quad n\in\{1, \ldots, N\}
\end{equation}
\begin{equation}
    Z'_n = [z'_0, \ldots, z'_N]    
\end{equation}
\begin{equation}
    o = s'^{T}Z'
\end{equation} 
where $o$ is the output and $W_Q, W_K\ and\ W_V$ are weight matrices for query, key and value respectively, and N is the number of side modalities.

To incorporate a new modality into the model, there are two additions, a new cross-attention adapter and a new concatenation to $s$ and $Z'_n$ vectors above. One disadvantage of this is that model size keeps increasing with more modalities. To alleviate this, the cross-attention module proposed by~\cite{wacv} is used in place of the traditional one and results are shown in Section~\ref{section4}. It makes adding new modalities hard, but it is a trade-off between flexibility and optimizing the usage of resources.

\section{Experiments}
\label{section4}

\subsection{Datasets}
To show robustness, we experiment using three datasets with different tasks and modalities. First, an egocentric \textbf{Epic-Kitchens-100}~\cite{EK100} consisting of three modalities RGB, optical-flow, and audio for actions related to human-object-interactions. Second, \textbf{MPIIGroupInteraction}~\cite{mpii} which is a body language dataset aiming at understanding human behavior in human-to-human interactions. For this dataset, we use the following modalities, RGB and audio. Finally, we have \textbf{UDIVA~v0.5}~\cite{udiva} which tackles the task of human personality analysis, using also different modalities such as RGB, transcript, and audio. Our approach shows effectiveness on all three tasks, proving our approach of bringing adapters mechanisms into vision problems to tackle all the challenges mentioned in the previous parts.

\textbf{Epic-Kitchens-100}~(EK-100) is a first-view and human-object interaction dataset. It contains 89,977 segments of fine-grained actions annotated from 700 long videos. Footage length amounts to 100 hours. It consists of a total of 97 verbs and 300 nouns, each action is a combination of a verb + noun and has a total of 3806 action classes.

\textbf{MPIIGroupInteraction} dataset~(MPIIGI) is 26 hours of spontaneous human behavior with 15 distinct body language classes. This dataset presents a novel set of actions which are challenging in computer vision and human-behavior understanding. It consists of body language behaviors such as gesturing, grooming, or fumbling.

\textbf{UDIVA~v0.5} dataset~(UDIVA) is 90.5 hours of dyadic interactions among 147 participants distributed in 188 sessions, recorded using multiple audiovisual and physiological sensors. But only half of the data has been released. UDIVA's main task is personality recognition. It has 5 main classes: Openness, Conscientiousness, Extroversion, Agreeableness, and Neuroticism (OCEAN).

\subsection{SOTA Comparison}
In this section we compare our results to the existing SOTA methods for each dataset and related PETL methods. The aim is to achieve similar performance to the methods we compare against while having considerably less trainable parameters.

\subsubsection{SOTA Comparison on EK-100 dataset}
1) Multimodal methods: Table~\ref{tab:sota_ek} shows the highest accuracy of M\&M Mix~\cite{1313} on EK-100 dataset~\cite{EK100}. M\&M Mix~\cite{1313} processes each of the three modalities using three branches of ViViT at different spatial resolutions using different sizes of input tubelets and different variants of ViViT. They use additional modules to share information across views and models for different modalities. One branch has more parameters than Video Swin-B, so the total number of parameters is more than three times the number of parameters of Video Swin-B. When taking Video Swin-B trained on Kinetics-400 as the backbone, we achieve performance comparable to the state-of-the-art (SOTA) (only 1.4\% worse) with a minuscule number of trained parameters~(more than 13 times less). Using a self-supervised trained backbone, CLIP, we achieve SOTA results (with the base variant of the backbone).

2) PETL methods: Table~\ref{tab:sota_ek} shows the comparison against SOTA PETL techniques, Dual-path adapters~\cite{dual_path} and ST-Adapters~\cite{st_adapter}. They do not provide these results and the results stated are from our own experiments using their code. We achieve considerably better performance when the pretraining protocol is the same. The results shows that it is hard to overcome the gap created by better pretraining as PETL techniques add minimal processing capacity. But, our plugins allow better performance when self-supervised pretrained backbones are not available.

To show the robustness of our proposed adapters design, we compare the proposed MHVA against the typical adapters from AdaptFormer~\cite{AF}. We compare the results of CM3T and adapters without additional modalities. Scaled parallel adapters (used in AdaptFormer) with PT achieve 28.7\% whereas MHVA achieves 39.8\%. This shows that our design of adapters is more robust. The motivation for the change discussed in the methodology section is thus justified from these results.

\begin{table}[ht]
\centering
\caption{SOTA comparison on EK-100. \textit{Acronyms- MHVA: Multi-head vision adapter, PT: Prefix Tuning, CAA: Cross Attention Adapters, CM3T: MHVA + PT + CAA, K400: Kinetics-400. Epochs presented are the number of epochs taken for convergence. All backbones other than CLIP-B are pretrained on Kinetics-400.}}
\label{tab:sota_ek}
\resizebox{\columnwidth}{!}{
\begin{tabular}{|l|c|c|c|c|}
\hline
Method & Backbone & Top-1 accuracy (\%) & Epochs &GFPLOs \\\hline
\multicolumn{2}{c}{Multimodal methods} \\ \hline
M\&M Mix~\cite{1313} & ViViT & 49.6 & 50 & \textgreater4790 \\\hline
CM3T & Video Swin-B & 48.2 & 22 & 616\\
CM3T & CLIP-B & \textbf{50.1} & 18 & 754\\ \hline
\multicolumn{2}{c}{PETL Methods} \\ \hline
Dual-Path Adapters\cite{dual_path} & ViT-B & 35.8 & 21 & 642\\
ST-Adapters\cite{st_adapter} & ViT-B & 34.3 & 18 & 911\\
Dual-Path Adapters\cite{dual_path} & CLIP-B & {44.8} & 24 & 642\\
ST-Adapters\cite{st_adapter} & CLIP-B & 44.1 & 18 & 911\\
Adaptformer\cite{AF} + PT & Video Swin-B & 28.7 & 6 & 357 \\\hline
MHVA + PT & Video Swin-B & 39.8 & 14 & 449\\
MHVA + PT & CLIP-B & \textbf{45.5} & 13 & 589 \\\hline
\end{tabular}
}
\end{table}

\subsubsection{SOTA Comparison on UDIVA and MPIIGI}
For UDIVA and MPIIGI, we compare to FAt transformers~\cite{wacv}, the SOTA for these datasets. FAt transformers have a lot of additions, specifically for UDIVA, which is the reason for their good performance. They have additional input branches with face crops and contextual videos and a complex method for preprocessing too. Tables~\ref{tab:sota_udiva} and~\ref{tab:sota_mpii} show a comparison against the published results. As for MPIIGI, we achieve better results with transfer learning techniques than FAt transformers which are fully finetuned. There are two reasons for this. One is that MPIIGI is a small dataset and it is easier for these PETL techniques to converge. The second reason is that Kinetics-400 is very close to MPIIGI and the CM3T backbone networks are initialized very well. This enables adapters to work better. In summary, CM3T achieves results equivalent to the SOTA for these two datasets with around 5 times less trainable parameters as compared to the previous SOTA. Using CLIP backbone, we achieve SOTA results.

We also show that our findings are consistent in the domain of PETL methods as we outperform ST-Adapters using our plugins.

\begin{table}[ht]
\centering
\caption{SOTA comparison on UDIVA. Acronyms from Table~\ref{tab:sota_ek}.}
\label{tab:sota_udiva}
\resizebox{0.75\columnwidth}{!}{
\begin{tabular}{|l|c|c|c|}
\hline
Method & Backbone & Mean MSE & Epochs \\ \hline
 \multicolumn{2}{c}{Multimodal methods} \\ \hline
FAt transformers\cite{wacv} & - & 0.72 & 30 \\\hline
CM3T & Video Swin-B & {0.69} & 27 \\
CM3T & CLIP & \textbf{0.65} & 22\\ \hline
\multicolumn{2}{c}{PETL Methods} \\ \hline
ST-Adapters\cite{st_adapter} & CLIP & 0.91 & 14 \\
MHVA + PT & CLIP-B & \textbf{0.8} & 14 \\ \hline
\end{tabular}
}
\end{table}

\begin{table}[ht]
\centering
\caption{SOTA comparison on MPIIGI. Acronyms from Table~\ref{tab:sota_ek}.}
\label{tab:sota_mpii}
\resizebox{0.7\columnwidth}{!}{
\begin{tabular}{|l|c|c|c|}
\hline
Method & Backbone & mAP & Epochs \\ \hline
 \multicolumn{2}{c}{Multimodal methods} \\ \hline
FAt transformers\cite{wacv} & - & 0.899 & 18 \\ 
CM3T & Video Swin-B & {0.901} & \hphantom{0}9 \\ 
CM3T & CLIP & \textbf{0.918} & 11\\ \hline
\multicolumn{2}{c}{PETL Methods} \\ \hline
ST-Adapters\cite{st_adapter} & CLIP & 0.886 & 14 \\
MHVA + PT & CLIP-B & 0.\textbf{894} & 9 \\ \hline
\end{tabular}
}
\end{table}

\subsubsection{Baseline Comparison}
% Here we can say that its because these transformers are pretrained on datasets like kinetics can extract good features hence the added few paramters can adapt to the new task easily.
Video Swin is one of the SOTA transformers trained on many datasets and tasks, hence it is chosen as the backbone. In Table~\ref{tab:baseline} we compare to full-finetuning the backbone vs. frozen backbone and only our plugins trained. For each dataset, we compare for multimodal input and only RGB input. 
% Note that we add cross-attention adapters~(CAA) to both, fully finetuned Video Swin-B and CM3T, to have a fair comparison.

For only RGB input, we achieve slightly lower results than full-finetuning. This is in tune with what we expect as traditionally pretrained backbones do not provide good generalizable features that can extend to other datasets and adapters have a limited capacity to take into account the distribution shift of the input. But as shown for EK-100, our plugins perform better than SOTA PETL methods when the same pretraining is applied.

Looking at multimodal input, for EK-100 dataset, CM3T achieves an accuracy of only 0.7\% lower than the fully finetuned model. Our method achieves comparable results with only 22.3\% parameters whereas Video Swin-B combining CAA goes up to 109.5\% parameters~(compared to Video Swin-B). Top-1 accuracy is the metric used here.

Moreover, for UDIVA~\cite{udiva} and MPIIGI~\cite{mpii}, we achieve the same results with CM3T as with full-finetuning and again with only 22.3\% of the total number of parameters in Video Swin-B. Mean MSE and mAP are used as metrics for them respectively.

\begin{table}[h]
\centering
\caption{Baseline comparison. Acronyms from Table~\ref{tab:sota_ek}. Top-1 accuracy for EK-100, MSE for UDIVA and mAP for MPIIGI. Number of trained parameters are reported on a relative scale, 100\% is equivalent to 88M.}
\label{tab:baseline}
\resizebox{\columnwidth}{!}{
\begin{tabular}{|l|c|c|c|c|c|}
\hline
Method & Backbone & Eval. metric & Epochs & Trained params \\\hline
\multicolumn{5}{c}{RGB input (EK-100)} \\ \hline
Full finetuning & Video Swin-B & \textbf{41.7\%} & 49 & 100.0\% \\
MHVA + PT & Video Swin-B & 39.8\% & 14 & 12.8\% \\ \hline
\multicolumn{5}{c}{Multimodal input (EK-100)} \\ \hline
Full finetuning + CAA & Video Swin-B & \textbf{48.9\%} & 56 & 109.5\% \\
CM3T & Video Swin-B & 48.2\% & 22 & 22.3\% \\\hline

\multicolumn{5}{c}{RGB input (UDIVA)} \\ \hline
Full finetuning & Video Swin-B & \textbf{0.82} & 51 & 100.0\% \\
MHVA + PT & Video Swin-B & 0.85 & 35 & 12.8\% \\ \hline
\multicolumn{5}{c}{Multimodal input (UDIVA)} \\ \hline
Full finetuning + CAA & Video Swin-B & {0.69} & 32 & 116.1\% \\
CM3T & Video Swin-B & {0.69} & 27 & 28.9\% \\\hline

\multicolumn{5}{c}{RGB input (MPIIGI)} \\ \hline
Full finetuning & Video Swin-B & \textbf{0.887} & 17 & 100.0\% \\
MHVA + PT & Video Swin-B & 0.882 & 8 & 12.8\% \\ \hline
\multicolumn{5}{c}{Multimodal input (MPIIGI)} \\ \hline
Full finetuning + CAA & Video Swin-B & 0.901 & 18 & 116.1\% \\
CM3T & Video Swin-B & {0.901} & 9 & 28.9\% \\\hline
\end{tabular}
}
\end{table}

\subsubsection{Cross-Attention Module}
An interesting thing to note is that MTV-B which is the base model for M\&M Mix and uses only RGB videos as input, achieves 46.7\% accuracy and there is only a 2.9\% accuracy increase when optical flow and audio are added to it. We achieve a higher increase of 8.4\% with CM3T when the two modalities are added. This might be because MTV-B is a better backbone as compared to Video Swin-B and captures most of the information present in optical flow already as optical flow is also a visual feature. Thus adding optical flow does not increase performance for them as much as us. This proves the efficacy of cross-attention adapters as we achieve similar performance to M\&M Mix, even when we are comparatively farther as compared to MTV-B. Moreover, we compare two methods for cross-attention: MMCA~\cite{wacv} and our proposed CAA and we observe that with our proposed solution we can achieve 0.5\% higher accuracy, showcasing robustness and efficacy of the proposed CAA. All results are in Table~\ref{tab:CAA}.

\begin{table}[ht]
\centering
\caption{Experiments for efficacy of CAA. Acronyms from Table~\ref{tab:sota_ek}. \textit{MMCA: multimodality cross-attention~\cite{wacv}}}
\label{tab:CAA}
\resizebox{\columnwidth}{!}{
\begin{tabular}{|l|c|c|c|c|}
\hline
Method & Backbone & Accuracy (\%) & Epochs & Trained params \\ \hline
 \multicolumn{5}{c}{RGB input} \\ \hline
{MTV-B}\cite{MTV-B} & - & 46.7 & 80 & \textgreater100.0\% \\
MHVA + PT & Video Swin-B & 39.8 & 14 & \hphantom{\textgreater0}12.8\% \\ \hline
\multicolumn{5}{c}{Multimodal input} \\ \hline
M\&M Mix\cite{1313} & - & 49.6 & 50 & \textgreater300.0\% \\
CM3T: MHVA + PT + CAA & Video Swin-B & 48.2 & 22 & \hphantom{\textgreater0}22.3\% \\ 
MHVA + PT + (MMCA\cite{wacv}) & Video Swin-B & 47.7\% & 24 & \hphantom{\textgreater0}22.7\%\\ \hline
\end{tabular}
}
\end{table}

\subsection{MHVA / PT}
MHVA and PT work well, as shown above. But, PT alone does not work very well as it tries to find learnable fixed inputs to be added to the actual input to provide context, but since supervised pretrained models do not give good relevant features for a different dataset, these inputs are not very useful unless combined with MHVA which provides a way for the model to learn the distribution shift in the input associated with the new dataset.

\subsection{Different Backbones}
% Table~\ref{tab:vivit-backbone} proves one of our previous claims that our method can be implemented with any backbone. In this study, we implement our proposed solution with a ViViT-B and we observe the same pattern in the results as with the Video Swin-B. 
This experiment supports our claim of CM3T being model-agnostic. Our plugins trained along with frozen ViViT-B achieve even better performance than full-finetuning. Table~\ref{tab:vivit-backbone} shows CM3T achieving better results with Video Swin as it is a better backbone, but this comparison does not say anything about out modules and is included here just for completeness.

\begin{table}[ht]
\centering
\caption{Results using different backbones. Experiments were done on EK-100 dataset. Acronyms from Table~\ref{tab:sota_ek}.}
\label{tab:vivit-backbone}
\resizebox{0.7\columnwidth}{!}{
\begin{tabular}{|l|c|c|}
\hline
Method & Backbone & Accuracy (\%) \\ \hline
\multicolumn{3}{c}{\textit{Backbone with supervised pretraining using K400}} \\ \hline
Full finetuning & ViViT-B & 37.4\% \\ 
MHVA + PT & ViViT-B & 38.1\% \\ \hline
CM3T & ViViT-B & 44.3\% \\ \hline
\end{tabular}
}
\end{table}

% \begin{table}[ht]
% \centering
% \caption{Ablation study on different component of our proposed architecture on EK100 dataset.}
% \label{tab:ablation_cm3t}
% \resizebox{\columnwidth}{!}{
% \begin{tabular}{|l|c|}
% \hline
% & \textbf{Top-1 Accuracy} \\ \hline
% Video Swin-B (frozen) + MHVA & 36.8 \\ \hline
% Video Swin-B (frozen) + PT & 23.3 \\ \hline
% Video Swin-B (frozen) + MHVA + PT & 39.8 \\ \hline
% Video Swin-B (frozen) + MHVA + PT + CAA & \textbf{48.2} \\ \hline
% \end{tabular}
% }
% \end{table}

%Table~\ref{tab:ablation_cm3t} proves that all the component in our architecture are necessary and complete each other as we achieve our best results when combining the multi-head adapters with the prefix tuning, moreover our multimodality cross-attention adapters (CAA), proves to be very effective as it improves the results by 8.4\%, hence it can capture more salient features and dependencies between different modalities compared to existing approaches such as in M\&M or FAt transformers.

%\subsection{Ablation studies}
%In this section we are going to look into more details of how different component contributes in our proposed architecture.

\subsubsection{Computational Resources}
We state that CM3T saves computational resources and we have already discussed a reduction in trainable parameters. Table~\ref{tab:sota_ek} shows that fewer epochs are required for the convergence of models with our plugins and also low FLOPs. For just finetuning RGB models, multi-head vision adapters and prefix tuning require a third of the time as compared to full finetuning. For adding a new modality, given an embedding corresponding to features of the new modality, only 5.8M additional parameters are required (with Video Swin-B as the backbone).

\section{Conclusion}
In this work, we presented CM3T (Cross Multimodal Multi-dataset Multitask Transformer), a framework for using common pretrained video classification models with a transformer-based architecture. The framework consists of three modules, two introduced by us, multi-head vision adapters and cross-attention adapters, and one already existing, prefix tuning. We show that in contrast to previous related works, these work well without specific pretraining or training methods (self-supervised methods) and study different variants. This work helps bridge the gap between research and practical applications of video classification models by making it easier to adapt existing work to new datasets and tasks, and also to utilize additional available modalities. Also, the framework benefits from weight sharing across different datasets for the same  modalities.

The limitation of this approach is that if the dataset used for pretraining is very dissimilar to the target one, the results will not be good. The frozen pretrained model needs to have the relevant information for the target task or dataset. Using various data augmentation, self-learning methods, or fully finetuned smaller models might give better results. For future work, combining adapters with selective finetuning of the model might resolve the above issue while keeping a low number of trainable parameters.

\section*{Ethical Statement}
This research adheres to the highest ethical standards, ensuring the welfare, dignity, and rights of all individuals involved. Even though this work does not record new data, the used data was collected with an informed consent obtained from all participants. No harm or bias has been detected during our research activities. We contribute to the advancement of knowledge while prioritizing the well-being of those involved within the scope of the European GDPR regulations.

\section*{Acknowledgements}
This work was supported by European Union under Horizon Europe project GAIN (GA \#101078950) and by the French National Research Agency under UCA\textsuperscript{JEDI} Investments into the Future (ANR-15-IDEX-01).

%%%%%%%%% REFERENCES
{\small
\bibliographystyle{ieee_fullname}
\bibliography{egbib}
}

\end{document}

%%%%%%%%% TITLE - PLEASE UPDATE
\title{CM3T: Framework for Efficient Multimodal Learning for Inhomogenous Interaction Datasets - Supplementary Material}

% \author{First Author\\
% Institution1\\
% Institution1 address\\
% {\tt\small firstauthor@i1.org}
% % For a paper whose authors are all at the same institution,
% % omit the following lines up until the closing ``}''.
% % Additional authors and addresses can be added with ``\and'',
% % just like the second author.
% % To save space, use either the email address or home page, not both
% \and
% Second Author\\
% Institution2\\
% First line of institution2 address\\
% {\tt\small secondauthor@i2.org}
% }
\maketitle

\section{Training Details}
For multi-head vision transformers, the bottleneck dimension used is $1/4$ multiplied by the channel dimension of the embedding for the respective block of the Video Swin-B. A smaller dimension size produces worse results, and a larger size produces similar results. For EK100, we use a slightly larger bottleneck dimension ($3/8$ times in place of $1/4$) for the last block of the Video Swin-B. For prefix tuning, the prefix channel dimension used is a minimum of $64$ and $1/8$ multiplied by the channel dimension of the embedding for the respective block of the Video Swin-B. The bottleneck dimension for the generation of the prefixes is $1/4$ multiplied by the channel dimension of the prefix. Smaller prefixes provide worse results. Larger prefixes make the networks overfit very fast. The model starts overfitting after 5-6 epochs with a big prefix. Prefix tuning is a shortcut for the network to force attention to focus on particular features by learning fixed additional inputs to keys and values. This allows it to easily learn patterns in the inputs or activations of the training set and thus overfit.

For cross-attention, we use feature embeddings extracted from side modalities. We use trill-distilled~\cite{trill} for obtaining audio features. Roughly 15 time steps of the audio features correspond to 128 frames in the videos (we use a stride of 4 frames for our input and Video Swin-B takes 32 frames as input). We use TV-L1 optical flow estimation and bninception~\cite{bninception} is used for its feature extraction. The features corresponding to each frame in the RGB video are taken, so the input of the temporal dimension is the same as RGB videos. Conv-1D is used for temporal pooling of all side modalities as Video Swin-B divides each input embedding into voxels and applies attention to each of them homogeneously. For simplicity, we give context from the side modalities for the whole input to every individual voxel. Also, we use performers~\cite{performer} for cross-attention to make it more efficient.

The cross-attention adapters are added to the first two and last two layers of each block of Video Swin-B. For the last layer of the last block, we use traditional cross-attention by changing the input for the value to be the same as the key and use the cross-attention adapters for late fusion in place of modifying attention. This provides slightly better results as the information from the other modalities passes further along with the value.

For training, we use 8/16 Tesla V100 GPUs with a batch size of 3 per GPU for adapters and 2 for full finetuning. These are the largest batch sizes we can fit on one GPU for each case. We train for varying numbers of epochs, stopping if performance does not increase for 6 epochs. The learning rate is modified according to the batch size. Video Swin transformers use a batch size of 8 per GPU and a starting learning rate of 0.0003. CM3T uses 0.0015 for batch size of 2 and 0.0018 for batch size 3. The weight decay is 0.05. Weight initialization for downscaling weights is used as Kaiming initialization, zero initialization for upscaling. Rest weight initializations are either from the pretrained model or default initialization from PyTorch. We use Video Swin-B pretrained on SSv2 dataset for experiments on EK100 dataset. For the experiments on the other datasets, we use the same model pretrained on Kinetics400 dataset. SSv2 is an egocentric dataset, similar to EK100 and pretrained Video Swin-B uses a larger window size for it, so we chose it for EK100. Kinetics400 is similar to the other datasets, so we use it for experiments on the others.

\section{Cross-attention adapters' behaviour with different modalities at different levels} \label{secA}
We apply cross-attention to the first and last two layers of each block in video swin transformers \cite{swin} (each block has multiple stacked transformer encoders and each block has different spatial resolutions for the patches being processed). If the blocks have only two layers, we apply them to both the layers. In this subsection, we study the importance of cross-attention at different levels for different modalities, by removing adapters from different blocks. Table \ref{tableA1} shows the results. This can be used to prune the architecture for specific modalities. We see that for audio and transcript, later layers are more important, whereas for optical flow, earlier layers are more important. Block 3 is the biggest block and is needed for good results for all side modalities.

\begin{table}
\begin{center}
\caption{Results for ablation study in section \ref{secA}. The entries show cross attention removed from a particular block, this is represented by "-" in  the table. For example, "CM3T - Block 1" represents CM3T without cross-attention in  Block 1. Also, block 1 is closest to input and block 4 is the last block before classification head.}
\label{tableA1}
\begin{tabular}{|l|c|}
\hline
Method & Performance \\
\hline
\multicolumn{2}{c}{Audio(MSE)}\\
\hline
UDIVA(CM3T) & 0.69\\
UDIVA(CM3T - Block 1) & 0.72\\
UDIVA(CM3T - Block 2) & 0.73\\
UDIVA(CM3T - Block 3) & 0.81\\
UDIVA(CM3T - Block 4) & 0.78\\
\hline
\multicolumn{2}{c}{Audio(Top-1 Accuracy)}\\
\hline
EK100(CM3T) & 48.2\%\\
EK100(CM3T - Block 1) & 47.8\%\\
EK100(CM3T - Block 2) & 47.5\%\\
EK100(CM3T - Block 3) & 46.4\%\\
EK100(CM3T - Block 4) & 47.1\%\\
\hline
\multicolumn{2}{c}{Transcript(MSE)}\\
\hline
UDIVA(CM3T) & 0.69\\
UDIVA(CM3T - Block 1) & 0.70\\
UDIVA(CM3T - Block 2) & 0.73\\
UDIVA(CM3T - Block 3) & 0.82\\
UDIVA(CM3T - Block 4) & 0.79\\
\hline
\multicolumn{2}{c}{Optical Flow(Top-1 Accuracy)}\\
\hline
EK100(CM3T) & 48.2\%\\
EK100(CM3T - Block 1) & 45.3\%\\
EK100(CM3T - Block 2) & 45.8\%\\
EK100(CM3T - Block 3) & 44.2\%\\
EK100(CM3T - Block 4) & 46.6\%\\
\hline
\end{tabular}
\end{center}
\end{table}

\section{Adding adapters to cross-attention adapters}
Since modalities are repeated across tasks and datasets, we see that training the entire cross-attention adapter module is not necessary. We can simply add scalable parallel adapters to the cross-attention modules. For this, the initial embedding is directly taken from the pretrained model and not the multi-head vision adapters, rest stays the same. Table~\ref{tableB1} shows the results for this experiment. We train cross-attention adapters for audio using EK100 and show results for UDIVA with normal cross-attention adapters and adapters added to cross-attention adapters.

\begin{table} 
\begin{center}
\caption{Result for adding adapters to cross-attention adapters.}
\label{tableB1}
% \label{tableA1}
\begin{tabular}{|l|c|}
\hline
Method & Performance \\
\hline
\multicolumn{2}{c}{UDIVA(MSE)}\\
\hline
CM3T & 0.690\\
CM3T (with adapters added to CA) & 0.689\\
\hline
\end{tabular}
\end{center}
\end{table}

\section{Ablation study: Our additions compared to adapters in Adaptformer}
Table~\ref{tableC1} shows the results of this ablation study.

\begin{table} 
\begin{center}
\caption{Ablation study results. $a_0$ represents trainable scaling factor. $a_1$ represents changing activation function to ReLU.}
\label{tableC1}
\begin{tabular}{|l|c|}
\hline
Method & Performance \\
\hline
\multicolumn{2}{c}{EK-100 (Accuracy(\%))}\\
\hline
Scaled parallel adapters + PT & 28.7\\
MHVA + PT & 39.8\\
MHVA - $a_0$ + PT & 38.7\\
MHVA - $a_1$ + PT & 39.1\\
\hline
\end{tabular}
\end{center}
\end{table}

%%%%%%%%% REFERENCES
{\small
\bibliographystyle{ieee_fullname}
\bibliography{egbib}
}